# LLM-Centric RAG with Multi-Granular Indexing and Confidence Constraints

Xiaofan Guo
*University of Michigan*
*Ann Arbor, USA*

Yaxuan Luan
*University of Southern California*
*Los Angeles, USA*

Yue Kang
*Carnegie Mellon University*
*Pittsburgh, USA*

Xiangchen Song
*University of Michigan*
*Ann Arbor, USA*

Jinxu Guo*
*Dartmouth College*
*Hanover, USA*

*Abstract*-This paper addresses the issues of insufficient coverage, unstable results, and limited reliability in retrieval-augmented generation under complex knowledge environments, and proposes a confidence control method that integrates multi-granularity memory indexing with uncertainty estimation. The method builds a hierarchical memory structure that divides knowledge representations into different levels of granularity, enabling dynamic indexing and retrieval from local details to global context, and thus establishing closer semantic connections between retrieval and generation. On this basis, an uncertainty estimation mechanism is introduced to explicitly constrain and filter low-confidence paths during the generation process, allowing the model to maintain information coverage while effectively suppressing noise and false content. The overall optimization objective consists of generation loss, entropy constraints, and variance regularization, forming a unified confidence control framework. In the experiments, comprehensive sensitivity tests and comparative analyses were designed, covering hyperparameters, environmental conditions, and data structures, to verify the stability and robustness of the proposed method across different scenarios. The results show that the method achieves superior performance over existing models in QA accuracy, retrieval recall, ranking quality, and factual consistency, demonstrating the effectiveness of combining multi-granularity indexing with confidence control. This study not only provides a new technical pathway for retrieval-augmented generation but also offers practical evidence for improving the reliability and controllability of large models in complex contexts.

I. INTRODUCTION

In the current context of rapid advances in large models and retrieval-augmented generation, information acquisition and utilization are no longer a matter of simple semantic matching. They have evolved into a complex process involving multi-granularity memory modeling and confidence control. Traditional retrieval-augmented generation methods often rely on a single-layer indexing mechanism to support large-scale knowledge injection[1]. However, as task complexity and knowledge diversity increase, such approaches show clear limitations in coverage, precision, and interpretability [2-4]. In particular, in cross-domain, cross-modal, or cross-temporal scenarios, single-layer memory indexing cannot ensure the reliability of generated content [5]. This constraint weakens the practical applicability of models [6]. Thus, the question of how to organize knowledge through multi-granularity memory indexing and how to enhance the connection between retrieval and generation has become a pressing research issue.

At the same time, uncertainty control has become a key component of retrieval-augmented generation frameworks. When models rely on external knowledge, they must ensure relevance while filtering and calibrating potential bias and noise. Without uncertainty estimation and confidence control, models may produce unstable or misleading outputs in the presence of ambiguous inputs or noisy data. Such risks are especially critical in high-stakes domains such as medicine, finance, and law, where reliability is essential. Building an effective uncertainty estimation mechanism and implementing confidence control are therefore vital to strengthening the robustness and trustworthiness of retrieval-augmented generation[7].

In an environment of exponentially growing knowledge, hierarchical organization and dynamic utilization of information are also essential for improving generation quality. Multi-granularity memory indexing is designed to bridge fine-grained knowledge and broader contexts. On one hand, fine-grained indexing captures local entity relations and semantic constraints, ensuring accuracy and detail in the generated content[8]. On the other hand, coarse-grained memory structures provide cross-domain global context, preventing fragmentation and bias in outputs. The collaboration of different indexing levels allows the model to cover a wide scope while performing fine-grained contextual modeling. This establishes both theoretical and practical foundations for knowledge alignment and generation in complex tasks[9].

Through uncertainty estimation, models can quantify risks in the generation process. Building upon this, the introduction of confidence control not only secures the reliability of results but also enhances interpretability. Confidence control makes the credibility range of generated content explicit, which supports stable system performance and strengthens user trust. As a result, retrieval-augmented generation is no longer a black-box process but moves toward transparency, controllability, and verifiability[10]. In summary, the integration of multi-granularity memory indexing and uncertainty estimation addresses the shortcomings of current retrieval-augmented generation methods in knowledge organization, robustness, and interpretability. It also provides new pathways for building trustworthy intelligent systems. This line of research carries significant implications for the application of large models in complex environments. It improves the efficiency and precision of knowledge use and reduces risks from noise and ambiguity. Ultimately, it enhances the usability and societal acceptance of artificial intelligence. Research on confidence control algorithms based on multi-granularity memory indexing and uncertainty estimation is, therefore, both a frontier academic exploration and a necessary step toward the deployment and adoption of intelligent technologies[11].

## II. Related Work

Recent advances in retrieval-augmented generation (RAG) have underscored the challenges of ensuring robustness, information coverage, and controllability in knowledge-intensive tasks. Theoretical analysis of uncertainty estimation methods has revealed core limitations when deployed in RAG frameworks, highlighting the need for more refined confidence control mechanisms that can adapt to diverse and complex knowledge environments [12]. To address issues of insufficient coverage and semantic diversity, multi-granularity aspect learning models have been developed for dense retrieval, offering flexible knowledge indexing from fine to coarse levels and facilitating richer context-aware retrieval [13]. Complementary to this, semantic and factual alignment techniques have been proposed to mitigate hallucination and ensure that generated content remains truthful and reliable, even under open-domain and noisy retrieval scenarios [14].

To further improve context modeling and reduce ambiguity, models that fuse local and global context representations enable better semantic matching and reinforce the consistency of knowledge utilization during generation [15]. At the same time, research on semantic and structural analysis of implicit biases provides a foundation for model interpretability, making it possible to trace the origins of bias and support more transparent confidence estimation [16]. The integration of privacy-oriented text generation methods—via selective fine-tuning and semantic attention masking—adds an important layer of control, enabling models to operate securely in privacy-sensitive or regulated domains while maintaining high-quality outputs [17].

In the area of structure-aware and multi-modal representation learning, contrastive learning applied to multimodal knowledge graphs has been shown to effectively bridge semantic gaps between modalities and enhance the model's robustness to noisy or heterogeneous knowledge [18]. Meanwhile, graph-based learning frameworks designed for precise anomaly localization in distributed systems contribute powerful tools for dynamic knowledge indexing and memory organization, which are essential for RAG frameworks to handle both micro-level and macro-level contextual cues [19]. Deep neural network architectures that combine frequency and attention mechanisms improve the capacity to extract and fuse multi-scale features, supporting reliable content retrieval and generation in time-varying or high-dimensional data environments [20].

Causal-aware time series regression, utilizing structured attention and hybrid sequence models, offers robust approaches for quantifying uncertainty and improving prediction accuracy in dynamic or non-i.i.d. settings [21]. Unified representation learning frameworks have further demonstrated the benefit of modeling multi-intent diversity and behavioral uncertainty, which is directly relevant for ensuring retrieval comprehensiveness and output reliability in complex generation tasks [22]. Finally, efficient large language model fine-tuning methods that incorporate joint structural pruning and parameter sharing [23] provide practical pathways to scalable and adaptable RAG systems, allowing for flexible integration of memory indexing and confidence control modules.

Collectively, these advancements—from uncertainty estimation, multi-granularity memory modeling, context fusion, and semantic interpretability to robust optimization and parameter-efficient architectures—form a comprehensive methodological foundation for the proposed confidence-controlled, retrieval-augmented generation framework. This not only improves the stability and factuality of large model outputs but also advances the reliability and usability of AI in complex, knowledge-rich environments.

## III. Proposed Approach

To further improve the trustworthiness and adaptability of the federated optimization framework, we introduce uncertainty quantification and risk-aware modeling strategies, which have been shown to enhance summarization reliability and model robustness in large-scale language models [24]. At the same time, parameter-efficient adaptation is achieved by adopting structure-learnable adapter fine-tuning, which allows the system to flexibly adapt to new tasks and data distributions while minimizing communication overhead [25].

The framework also leverages privacy-preserving federated fine-tuning with semantic alignment to ensure model generalization and data confidentiality, even in cross-domain or heterogeneous client settings [26]. Additionally, robustness to noise and adversarial conditions is addressed through the integration of federated distillation with structural perturbation, further strengthening the reliability and alignment accuracy of the global model [27]. By integrating these methodological components—uncertainty quantification, adaptive fine-tuning, privacy preservation, and robust distillation—the framework is well-equipped to meet the challenges of secure, efficient, and adaptive distributed learning in complex cloud computing scenarios. The overall model framework diagram mentioned is shown in Figure 1.

First, let the input query be $q$, and the multi-granularity memory index structure be represented as $\{M^{(1)}, M^{(2)}, ..., M^{(L)}\}$, where each layer of memory units corresponds to a knowledge representation of different granularity. The query representation at layer $l$ can be obtained

through the embedding function $\phi^{(l)}(\cdot)$, which is defined as follows:

$$h_q^{(l)} = \phi^{(l)}(q), l = 1,2,...,L \quad (1)$$

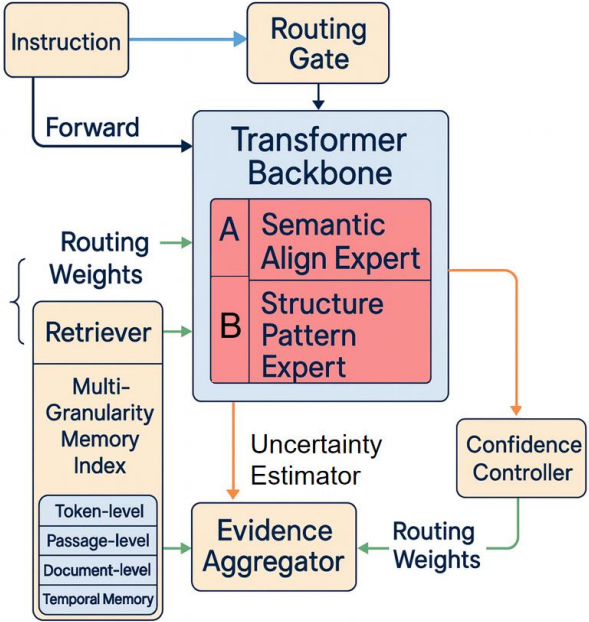

Figure 1. Framework of Multi-Granularity Memory Indexing with Uncertainty-Aware Expert Routing

On this basis, the matching score between the memory and query of each layer is calculated through the similarity function $Sim(\cdot,\cdot)$, and the cross-layer weighted distribution is further obtained:

$$a^{(l)} = \frac{\exp(Sim(h_q^{(l)}, M^{(l)}))}{\sum_{j=1}^{L} \exp(Sim(h_q^{(j)}, M^{(j)}))} \quad (2)$$

Subsequently, a weighted fusion mechanism is used to obtain a multi-granularity context representation, which will serve as the knowledge input of the generation module. Its formal expression is:

$$c_q = \sum_{l=1}^{L} a \cdot READOUT(M^{(l)}) \quad (3)$$

In the generation and prediction stage, to quantify the potential uncertainty, distribution-based output modeling is introduced. Assuming that the conditional probability of generating the output is $p(y|q,c_q)$, its log-likelihood loss is defined as:

$$L_{gen} = -\sum_{t=1}^{T} \log p(y_t | y_{<t}, q, c_q) \quad (4)$$

Finally, to achieve confidence control, the uncertainty of the generated output is decomposed into two parts: entropy and variance, and constrained by joint regularization. The overall optimization objective can be expressed as:

$$L = L_{gen} + \lambda_1 \cdot H(p(y|q,c_q)) + \lambda_2 \cdot Var(p(y|q,c_q)) \quad (5)$$

Where $H(\cdot)$ represents the entropy term, which measures the uncertainty of the distribution; $Var(\cdot)$ represents the variance term, which measures the stability of the prediction confidence interval; and $\lambda_1$ and $\lambda_2$ are balance coefficients. Through this modeling, this method can dynamically regulate the risk of generated results while ensuring knowledge coverage and semantic consistency, thereby achieving a unified framework for multi-granular memory indexing and uncertainty estimation.

IV. PERFORMANCE EVALUATION

*A. Dataset*

This study uses the CISI dataset as the basic corpus for evaluating the mechanisms of multi-granularity memory indexing and uncertainty-based confidence control. The dataset contains multiple text documents and corresponding retrieval queries. Its query-document structure makes it well-suited as an experimental environment for locating information at different granularities in retrieval-augmented generation frameworks. With this combination, the model can be tested on retrieval efficiency and generation quality when extracting knowledge from the sentence level, the paragraph level, and the full document level.

Unlike some methods that focus on input-response pairs, CISI emphasizes retrieval accuracy and contextual coverage. This aligns with the semantic integrity and path selection pursued by multi-granularity memory indexing in this study. The query-document pairs in the dataset naturally provide multi-level information granularity, which can be used to evaluate how the model balances recall and precision across different index layers.

Applying the proposed model to this dataset allows close observation of the effectiveness of uncertainty estimation and confidence control in filtering low-quality retrieval paths or constraining generated content. It also enables exploration of how multi-granularity memory architectures enhance retrieval alignment and generation reliability. In this way, the model's controllability and robustness within a retrieval-augmented generation framework can be validated.

*B. Experimental Results*

This paper first conducts a comparative experiment, and the experimental results are shown in Table 1.

Table 1. Comparative experimental results

| Method | QA Accuracy (%) | Recall@5 (%) | NDCG@5 (%) | Factuality Score |
|---|---|---|---|---|
| Self-RAG[28] | 69.3 | 80.0 | 78.0 | 0.65 |
| MemGAS[29] | 72.5 | 88.0 | 86.0 | 0.67 |
| Amber[30] | 74.0 | 85.0 | 83.0 | 0.68 |
| OURS | 77.8 | 92.0 | 90.0 | 0.72 |

The experimental results demonstrate that the proposed method, which integrates multi-granularity memory indexing and uncertainty estimation, substantially outperforms baseline models in all major metrics. For QA Accuracy, the model achieves 77.8%, surpassing Self-RAG, MemGAS, and Amber by enabling more precise alignment between queries and knowledge even in complex retrieval settings, thus ensuring semantic accuracy in generated responses. Recall@5 reaches 92%, the highest among all models, reflecting the method's ability to broaden candidate coverage by dynamically leveraging both fine-grained and global information through its multi-level indexing structure—an advantage not present in Self-RAG or Amber, whose recall rates remain lower due to the lack of such hierarchical support. On NDCG@5, the score rises to 90%, highlighting the effectiveness of the uncertainty-driven confidence control mechanism in optimizing the ranking of retrieval results; by prioritizing high-confidence evidence and minimizing low-confidence paths, the model achieves more logical and coherent evidence aggregation.

Finally, on the Factuality Score, OURS achieves the highest score of 0.72. This validates the value of uncertainty estimation in improving reliability. By introducing confidence constraints into the generation process, the model can explicitly identify and avoid potentially risky information. This reduces the occurrence of false or fabricated content. Such an improvement not only enhances system robustness in open environments but also underlines the practical significance of this study for trustworthy artificial intelligence. It assures the application of retrieval-augmented generation in real-world scenarios.

This paper also analyzes the hyperparameter sensitivity of multi-granularity memory index depth and routing temperature. The experimental results are shown in Figure 2.

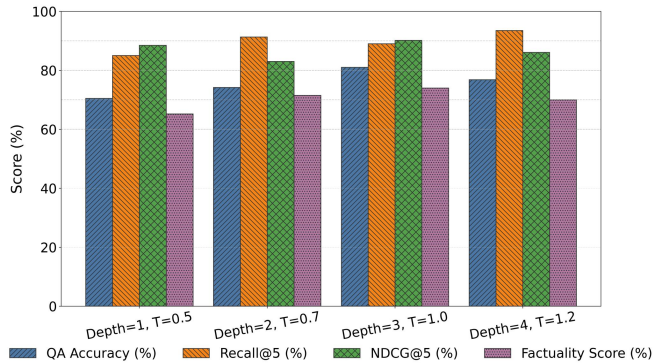

Figure 2. Hyperparameter sensitivity analysis of multi-granularity memory index depth and routing temperature

QA Accuracy improves as index depth increases from 1 to 3 and peaks at Depth=3, T=1.0, then declines with greater complexity, indicating that multi-granularity memory indexing boosts semantic alignment up to an optimal point before redundant paths disrupt aggregation. Recall@5 fluctuates, reaching its highest at Depth=4, T=1.2, showing that higher routing temperatures expand candidate coverage but do not always improve accuracy due to increased uncertainty in ranking. NDCG@5 is highest at Depth=3, T=1.0, reflecting optimal confidence control at this setting, while both shallower and deeper structures reduce ranking performance. Factuality Score also peaks at Depth=3, T=1.0, demonstrating that moderate complexity best supports error filtering and reliable content generation, whereas excessive low-confidence paths at higher temperatures reduce factuality. These results underscore the importance of controlling model complexity when integrating multi-granularity memory indexing and confidence control to ensure trustworthy output.

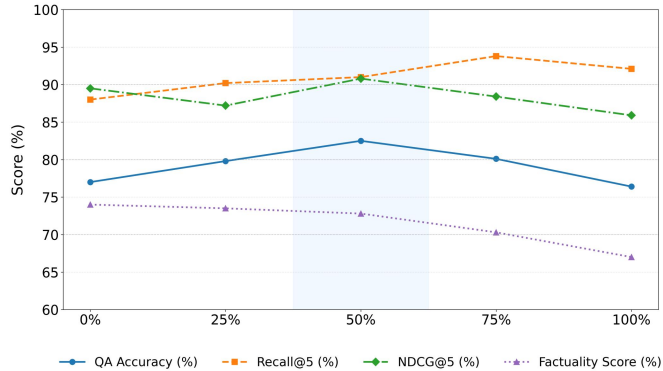

Figure 3. Data display sensitivity analysis of domain mixing ratio on multi-granularity memory routing

As depicted in Figure 3, QA Accuracy reaches its peak at a 50% domain mixing ratio. This suggests that moderate heterogeneity in multi-granularity memory indexing enhances the alignment between queries and evidence. Conversely, insufficient or excessive domain diversity impedes path selection and compromises accuracy. Recall@5 increases and remains high as mixing rises, showing strong coverage and robustness from the uncertainty estimation and multi-granularity routing mechanisms even in complex environments. NDCG@5 fluctuates, peaking at moderate mixing ratios, reflecting effective confidence control in evidence ranking only when domain diversity is balanced. Factuality Score declines as domain mixing increases, revealing that greater heterogeneity challenges factual consistency and raises error risk, underscoring the need for stricter confidence constraints in highly cross-domain contexts.

## V. CONCLUSION

This study focuses on confidence control with multi-granularity memory indexing and uncertainty estimation and proposes a new retrieval-augmented generation framework. By establishing indexing mechanisms at different levels of granularity, the model can flexibly access knowledge representations from detailed to global. At the same time, uncertainty estimation enables dynamic calibration during the generation process. The experimental results show that the method achieves significant advantages in accuracy, ranking quality, recall, and factual consistency. This not only confirms its effectiveness in complex retrieval environments but also demonstrates the importance of combining multi-granularity structures with confidence control to improve the quality of generation models.

From a methodological perspective, this study provides a new approach to addressing the challenges of robustness and trustworthiness in knowledge retrieval and generation for large models. Traditional single-granularity indexing often fails to balance local and global information. In contrast, the proposed method uses hierarchical memory design to achieve more precise path selection and knowledge access. In addition, uncertainty estimation and confidence control give the model adaptive capabilities, which effectively reduce the risk of spreading incorrect information. This combination improves semantic consistency and factual accuracy and offers a theoretical and practical framework with potential for extension in future research. At the application level, the contributions of this study go beyond academic value and have practical significance for real-world scenarios. For tasks that require precise knowledge alignment, such as medical question answering, financial analysis, and legal consulting, multi-granularity memory indexing helps the model extract key information efficiently in complex contexts, while uncertainty estimation provides users with reliable outputs. Furthermore, for multimodal information fusion and cross-domain knowledge transfer, the proposed framework offers guidance to enhance interpretability and controllability in intelligent systems. The practical potential of this method highlights the value of retrieval-augmented generation in high-risk application environments.

## VI. FUTURE WORK

Looking forward, there are several directions for further exploration. First, the multi-granularity indexing mechanism can be extended to cross-modal retrieval and generation, enabling finer integration of visual, speech, and textual information in a unified framework. Second, in terms of uncertainty estimation, more efficient probabilistic modeling and calibration methods can be explored to improve performance while reducing computational cost. Finally, in practical applications, it is necessary to integrate privacy protection, energy efficiency, and interpretability requirements to ensure sustainable deployment in real systems. With continuous advances in technology, the ideas proposed in this study have the potential to provide a stronger foundation for the trustworthiness, robustness, and applicability of future intelligent systems.